\pdfoutput=1

\documentclass[11pt]{article}

\usepackage{emnlp2021}

\usepackage{times}
\usepackage{latexsym}

\usepackage[T1]{fontenc}

\usepackage[utf8]{inputenc}

\usepackage{microtype}

\usepackage{float}
\usepackage{todonotes}
\usepackage{xspace} 
\usepackage[capitalize,noabbrev]{cleveref} 
\usepackage{multirow,booktabs, hhline}

\newcommand{\LEFTL}{Learning from Textbooks~(LEFT)\xspace}
\newcommand{\LEFTS}{LEFT\xspace}
\newcommand{\americangovernment}{American Government 2e\xspace}
\newcommand{\ushistory}{U.S. History\xspace}

\title{Perhaps PTLMs Should Go to School – \\ A Task to Assess Open Book and Closed Book QA}

\author{
    {\bf Manuel R. Ciosici}, {\bf Joe Cecil}, {\bf Alex Hedges},  {\bf Dong-Ho Lee}, \\ {\bf Marjorie Freedman}, {\bf Ralph Weischedel} \\
    \texttt{manuelc@isi.edu},~~\texttt{mrf@isi.edu},~~\texttt{weisched@isi.edu} \\
    Information Sciences Institute, University of Southern California
}

\begin{document}
\maketitle

\begin{abstract}
    Our goal is to deliver a new task and leaderboard to stimulate research on question answering and pre-trained language models (PTLMs) to understand a significant instructional document, e.g., an introductory college textbook or a manual. PTLMs have shown great success in many question-answering tasks, given significant supervised training, but much less so in zero-shot settings. We propose a new task that includes two college-level introductory texts in the social sciences (\americangovernment) and humanities (\ushistory), hundreds of true/false statements based on review questions written by the textbook authors, validation/development tests based on the first eight chapters of the textbooks, blind tests based on the remaining textbook chapters, and baseline results given state-of-the-art PTLMs.
    
    Since the questions are balanced, random performance should be \textasciitilde50\%. T5, fine-tuned with BoolQ achieves the same performance, suggesting that the textbook's content is not pre-represented in the PTLM. Taking the exam closed book, but having read the textbook (i.e., adding the textbook to T5's pre-training), yields at best minor improvement (56\%), suggesting that the PTLM may not have ``understood'' the textbook (or perhaps misunderstood the questions).  Performance is better (\textasciitilde60\%) when the exam is taken open-book (i.e., allowing the machine to automatically retrieve a paragraph and use it to answer the question).
    
\end{abstract}

\section{Introduction}
\label{sec:introduction}

Question answering (QA) is a yardstick for measuring machine understanding performance~\cite{Hermann_teaching_machines}. QA's popularity as an evaluation technique has led to several sub-categories: tasks can require a model to answer questions from either its background knowledge or from a short passage (e.g., SQuAD, \citealp{rajpurkar-etal-2016-squad}) or with information retrieval to allow the model to search for the answer in a large corpus (e.g., ARC, \citealp{bhakthavatsalam2021think}). Answering can take the form of true/false classification (BoolQ, \citealp{clarkBoolQExploringSurprising}), multiple-choice, span selection (SQuAD, \citealp{rajpurkar-etal-2016-squad}), or text generation (TriviaQA, \citealp{joshi-etal-2017-triviaqa}).

Transformer architectures optimized for speciﬁc QA formulations have driven recent progress in question answering. For example, some models target IR-oriented QA~\cite{guu2020realm} while others optimize their learning strategy to specific question types (e.g., by optimizing for expected answers to factoid questions, \citealp{roberts-etal-2020-much}). While specialization improves performance, it limits generalization. UniﬁedQA~\cite{khashabi-etal-2020-unifiedqa} takes a step forward by generalizing the architecture and training over multiple data sets with different QA formulations. 

Most research assumes that the information necessary to answer questions is either included with the query (e.g., BoolQ, SQuAD 1.1) or that the information was already stored in language models during initial pre-training or a task-specific second pre-training.\footnote{For example, \citet{roberts-etal-2020-much} adjust T5's masking strategy to target named entities as they expect named entities to be parts of answers.} However, this assumption limits language models relying on massive corpora~\cite{gaoPile800GBDataset,Raffel2019a} to learning oft-repeated facts \cite{petroni-etal-2019-language}. Valuable, domain-specific information seldom is repeated often enough to be captured by language models. An evaluation of domain-speciﬁc knowledge without access to a relevant text is even more challenging as simple strategies like identifying the answer by information retrieval are ineffective. Even reasoning tasks such as ARC~\cite{bhakthavatsalam2021think} only target general scientific knowledge and offer large text corpora to aid QA systems.

We propose \LEFTL, a new task to classify domain-specific statements drawn from a textbook's review questions as true or false using three evaluation configurations. The first configuration tests the ability to answer questions without any domain-specific material (e.g., applying a PTLM with no access to domain-specific knowledge). This setting is equivalent to a person taking the test before taking the class. In the second configuration, a model has access to the textbook's content and may encode the information in the textbook but may not access the textbook during the test; we call this \emph{closed book}. The second configuration tests a model's ability to learn by reading. In the third configuration, which we call \emph{open book}, models can access the textbook during the test. Thus, \LEFTS supports contrasting QA formulations and reading methods to explore the strengths and weaknesses of various QA approaches. The LEFT data and leaderboard are available at \url{https://leftleaderboard.isi.edu}.

\section{Related Work}
\label{sec:related_work}

\paragraph{Question Answering.}
Most previous research specializes QA models to target speciﬁc question formulations. Question answering with a relevant paragraph often relies on span selection~\cite{rajpurkar-etal-2016-squad,yang-etal-2015-wikiqa} or simple reasoning~\cite{clarkBoolQExploringSurprising}. Previous open-book QA methods first filter a large corpus to a small set of relevant documents using information retrieval ~\cite{karpukhin-etal-2020-dense,robertson2009probabilistic}. The document set then provides context for answering questions~\cite{dhingra2017quasar, dunn2017searchqa,joshi-etal-2017-triviaqa, nguyen2016ms}. Conversely, closed-book QA instead requires models to answer using only their implicit knowledge~\cite{roberts-etal-2020-much}. Taking a step towards generalizing QA, UnifiedQA~\cite{khashabi-etal-2020-unifiedqa} proposes a unified architecture that answers various question types relying partly on knowledge encoded in its language model.

\paragraph{Knowledge in Pre-trained Language Models.}
Pre-trained language models (PTLMs) have shown good performance in cloze-style queries~\cite{petroni-etal-2019-language}, fact-checking~\cite{thorne-etal-2018-fever}, entity linking~\cite{guo2018robust,hoffart-etal-2011-robust}, and open-domain QA~\cite{joshi-etal-2017-triviaqa, kwiatkowski-etal-2019-natural, petroni2020kilt}. However, in most cases, the PTLMs rely on knowledge learned from massive corpora during pre-training. \LEFTS tests domain-speciﬁc knowledge acquired from a textbook, a small corpus of only a few hundreds of thousands of words (see \Cref{tbl:data_overview}).

\paragraph{Textbook Question Answering.}
Researchers have explored machine understanding of elementary- and middle-school science textbooks by visual question answering~\cite{gomez-perez-ortega-2020-isaaq,Kembhavi2017tqa,kim-etal-2019-textbook} and information retrieval~\cite{bhakthavatsalam2021think}.
While existing textbook QA tasks focus on general knowledge (which can be gained by pre-training on general web corpora), \LEFTS focuses on domain-specific knowledge. Furthermore, it quantiﬁes pre-trained language models' pre-existing knowledge by requiring that models take the task \emph{before} and \emph{after} reading \LEFTS's two textbooks.

\section{Task Description}
\label{sec:task_description}

\begin{table}[t]
    \centering
    \scalebox{0.68}{
        \begin{tabular}{lrrrr}
            &
            \multicolumn{2}{c}{AG} & \multicolumn{2}{c}{USH} \\
            \cmidrule(lr){2-3} \cmidrule(lr){4-5} &
            Dev & Test & Dev & Test \\
            \midrule
            Num. chapters & 8 & 9 & 8 & 24 \\
            Text size (words) & 137\,620 & 138\,668 & 89\,765 & 301\,860 \\
            Num. statements & 186 & 214 & 274 & 412 \\
            \bottomrule
        \end{tabular}
    } 
    \caption{Data overview for the two textbooks: \americangovernment (AG) and \ushistory (USH).}
    \label{tbl:data_overview}
\end{table}

\LEFTL contains two machine-readable college-level introductory textbooks and a set of true/false statements manually derived from review questions written by the textbook authors. The task requires that systems based on language models classify the statements \emph{before} and \emph{after} reading the given textbook material to separate what was learned from the book from what was known before reading. ``Reading'' is any algorithm method that learns from the domain text without storing a copy of the text. To support comparisons with existing QA approaches, \LEFTS also supports the open-book setting, where a system can use a textbook paragraph when answering.

Our goal is to support testing pre-trained language models, e.g., T5~\cite{Raffel2019a}, and also those approaches that extract and store triples during reading (e.g., \emph{<U.S. Declaration of independence; signed; Aug 2, 1776>}). While learning corpora appear in other question answering tasks (e.g., ARC, 14M words, \citealp{bhakthavatsalam2021think}), the text included in \LEFTS is small and corresponds to the textbook chapters relevant to each question set. The largest text in \LEFTS contains only 300K words (for details, see \Cref{tbl:data_overview}).

\LEFTS includes two openly licensed\footnote{Both textbooks are licensed under the \href{https://creativecommons.org/licenses/by/4.0/}{Creative Commons Attribution License v4.0} license.} college-level introductory textbooks, \emph{\americangovernment}~\cite{krutz2019american} and \emph{\ushistory}~\cite{corbett2014us}, and true/false statements derived from each book's review questions. We manually rewrote each textbook's multiple-choice review questions into a balanced set of true and false statements.\footnote{We construct one true and one false statement for each question to obtain a balanced data set. For example, the question \emph{When was the U.S. Declaration of Independence signed? (A)(correct) August 2, 1776 (B) December 2, 1776, (C) August 2, 1746, (D) August 22, 1976} could become \emph{The U.S. Declaration of Independence was signed on August 2, 1776} (true) and \emph{The U.S. Declaration of Independence was signed on August 2, 1746} (false).}\textsuperscript{,}\footnote{For \ushistory's \emph{Dev} set, we also process questions written by a community of instructors.} We intentionally wrote the statements such that each \emph{true} and \emph{false} pair has high word overlap to deter classification strategies that rely on word overlap with the textbook. We include five sample statements from \LEFTS in \Cref{sec:sample_statements} and discuss statement correctness in \Cref{sec:statement_correctness}.

We measure task performance by \emph{accuracy}. Since the two textbooks are used in teaching college students, we do not release the correct labels (see the \nameref{sec:ethical_considerations} section). We split each textbook into a \emph{Dev} set consisting of the first eight chapters and a \emph{Test} set consisting of the remaining chapters (see \Cref{tbl:data_overview} for an overview). We allow unlimited submissions to the \emph{Dev} set, but for any submission, we only provide the overall \emph{accuracy} without feedback on which statements were correctly classified. This design decision aims to prevent divulging the correct answers (see the \nameref{sec:ethical_considerations} section).

\LEFTS has three evaluation configurations: \emph{(1) Prior-knowledge}; \emph{(2) Closed-book, after reading}; and \emph{(3) Open-book}. \textbf{Prior-knowledge} tests the ability to answer questions without any domain-speciﬁc material. Language models must rely solely on the knowledge learned from their large pre-training corpora. In the second configuration, \textbf{Closed-book, after reading}, models may access the textbook's content and may encode the information in the textbook but may not access the textbook during the test. For each set, models may read the set's corresponding textbook chapters, the entire textbook, or both textbooks. We require that all model submissions to this evaluation configuration also submit to \emph{Prior-knowledge}. Predictions \emph{before} reading (\emph{Prior-knowledge}) quantify the information included in each model through initial pre-training. The change in performance from \emph{Prior-knowledge} to \emph{Closed-book, after reading} illustrates each model's reading effectiveness. In the third configuration, \textbf{Open-book}, models can access the textbook or relevant chapter during the test. To support research on open-book question answering, with each statement, we include the textbook paragraph that provides the information necessary to classify the statement. In our experiments, we call this \emph{goldIR}. Thus, \LEFTS supports contrasting QA formulations and reading methods to explore the strengths and weaknesses of various QA approaches.

\section{Results}
\label{sec:results}

\begin{table*}[h]
    \centering
    \small
    \scalebox{1}{
            \begin{tabular}{lrrrr}
                &
                \multicolumn{2}{c}{\americangovernment} & \multicolumn{2}{c}{\ushistory} \\
                \cmidrule(lr){2-3} \cmidrule(lr){4-5} &
                Dev (186) & Test (214) & Dev (274) & Test (412) \\
                \midrule
                \textbf{Prior-knowledge} & & & \\
                T5-3B -ctx & 51.08 & 49.53 & 50.36 & 50.00\\
                GPT-Neo 2.7B -ctx & 52.69 & 48.13 & 51.09 & 49.27 \\
                \midrule
                \textbf{Closed-book, after reading} & & & \\
                T5 3B +pt -ctx & 56.45 & 52.34 & 50.73 & 50.00 \\
                GPT-Neo 2.7B +pt -ctx & 50.00 & 55.14 & 50.73 & 49.76 \\
                \midrule
                \textbf{Open-book} & & & \\
                T5-3B +ctx +sBERT & 60.22 & 61.21 & 55.47 & 59.95 \\
                T5-3B +pt +ctx +sBERT & 55.91 & 52.80 & 52.19 & 56.31\\
                T5-3B +ctx +goldIR & \textbf{71.51} & \textbf{74.30} & \textbf{68.61} & \textbf{68.69} \\
                T5-3B +pt +ctx +goldIR & 58.60 & 63.08 & 57.66 & 66.26 \\
                \bottomrule
            \end{tabular}
    } 
    \caption{Baseline accuracy with the current state-of-the-art language models. \ushistory's \emph{Dev} set consists of statements based on the textbook statements and on questions from a community of instructors. In the heading, each set's name is followed by its number of statements. The order of abbreviations reflects the order of operations. All models are fine-tuned with BoolQ; \emph{+/- ctx}~--~whether we included BoolQ's context during fine-tuning; \emph{+pt}~--~whether we pre-trained on the relevant textbook chapters.}
    \label{tbl:results}
\end{table*}

We illustrate baseline performance on \LEFTS using two state-of-the-art language models: T5~\cite{Raffel2019a} and GPT-Neo~(a GPT-3 architecture, \citealp{Brown2020}, trained on the open Pile corpus, \citealp{gaoPile800GBDataset}). We fine-tune the two language models using BoolQ~\cite{clarkBoolQExploringSurprising}. \Cref{tbl:results} shows results in \LEFTS's three evaluation settings: \emph{Prior-knowledge} (out-of-the-box language models fine-tuned on BoolQ), \emph{Closed-book, after reading} (language models with continued light pre-training on \LEFTS's text content), and \emph{Open-book} (where models have access to the relevant textbook paragraph). Since the \emph{Prior-knowledge} and \emph{Closed-book}  settings do not include the relevant paragraph for each question, we adjust fine-tuning to only use BoolQ's questions and ignore its text snippets. In the \emph{Open-book} setting, we consider automatically retrieved textbook paragraphs (using sBERT, \citealp{reimersSentenceBERTSentenceEmbeddings2019}) and manually identified the relevant paragraphs (gold information retrieval, goldIR). When selecting the relevant textbook content, we select one natural paragraph (i.e., as written by each textbook's authors). However, due to technical limitations imposed by T5's memory consumption, in our experiments, we limit the concatenated statements and paragraphs to a maximum length of 128 word pieces (see \Cref{sec:t5_hyperparameters}).

\subsection{Baseline Results}

T5 and GPT-Neo's scores are indistinguishable from the random baseline of 50\% in the \emph{Prior-knowledge setting}, suggesting that the textbooks query for information is either \emph{not} present in the two language models or not easily accessible. Continuing each model's pre-training with the relevant textbook parts sometimes helps, but not consistently. The lack of improvement \emph{after} reading is further evidence that the models memorize, but not in beneficial ways, i.e., they can complete sentences but do not learn the subject matter and cannot classify the statements, even after 20 epochs. It also suggests that the closed-book setting represents a new challenge for PTLMs.

Accuracy in the open-book setting is far higher, especially when using goldIR (i.e., a manually selected relevant paragraph). As in the closed book setting, we contrast models using only prior knowledge with models pre-trained on the textbook. Pre-training with the textbook never improves the system's accuracy, suggesting that even in this setting, the models are not learning by reading the textbook. The gap between goldIR- and sBERT-based retrieval suggests that there is room for retrieval-based improvement in the open-book setting. However, even with goldIR, T5 only achieves an accuracy of \textasciitilde70\%, suggesting that paragraph-based QA alone is not solved with existing models.

\section{Conclusions \& Future Work}
\label{sec:conclusions}

There are several natural directions in which we can extend and improve \LEFTS. We are extending \ushistory's \emph{Test} set as we did with the \emph{Dev} set by including statements based on questions written by a community of instructors. We are also collecting relevant paragraphs for the extra statements. Lastly, we are categorizing the kind of knowledge required to classify each statement to better understand what kinds of knowledge pose the most difficulties.

We draw several conclusions from this work. Foremost, \LEFTL represents a new type of challenge task for PTLMs, contrasted with the much-studied challenges of (1) common sense QA based on prior knowledge, (2) reading comprehension given a paragraph, and (3) QA using large domain-specific corpora, e.g., science at the elementary- or middle-school level. The task is intended to stimulate research on the following dimensions:
\begin{enumerate}
    \item Zero-shot learning, much as an entering college student could do when studying a textbook,
    \item Measuring a system's knowledge before vs. after ``reading'' the textbook,
    \item Capability in both closed-book and open-book question answering,
    \item The effect of IR accuracy on task accuracy compared to the system's language understanding performance.
\end{enumerate}
Our baseline studies show that T5 and GPT-Neo thus far are challenged to show improvement after reading the relevant textbook, that open-book evaluation is easier than closed-book (as it is for humans), and that the gating factor in \LEFTS is understanding the textbook and/or the question rather than paragraph retrieval.
The baseline results show there is much room for improvement.

\section*{Ethical Considerations}
\label{sec:ethical_considerations}
We have reflected on two ethical considerations when creating \LEFTL: content and environmental impact.

\textbf{Content.} The two textbooks in \LEFTS cover topics that include history, race, and politics. OpenStax textbooks follow a set of \emph{Diversity and Representation Development Guidelines}, which aim to ``properly represent genders, gender identities, races, cultures, geographies, ethnic backgrounds, disabilities, nationalities, ages, sexual orientations, socio-economic status, and diverse viewpoints''.\footnote{See \emph{Diversity and Representation Development Guidelines} in the instructor materials for each textbook.} As creators of an NLP task, we do not make any claims, nor do we comment on the topics covered in the two textbooks. Furthermore, we understand that documents as large and complex as textbooks are bound to contain inaccuracies. We invite users with specific content accuracy concerns to consult the official textbook errata included in each textbook's instructor resources.\footnote{See the \emph{Errata Release Notes} at \url{https://openstax.org/details/books/american-government-2e?Instructor\%20resources} for \emph{\americangovernment} and \url{https://openstax.org/details/books/us-history?Instructor\%20resources} for \emph{\ushistory}.}

Releasing labels for the statements in \LEFTS would indirectly reveal the correct answers for multiple-choice questions in the two textbooks. While both \americangovernment and \ushistory include answer keys, they are incomplete. We believe releasing the correct answers to all multiple-choice questions in the book would be detrimental to the intended primary users of the two textbooks; in other words, it might hinder students' learning. We only used full-time employees compensated according to U.S. law to rewrite the multiple-choice review questions in the two textbooks.

\textbf{Environmental.} We included baseline results based on large pre-trained language models. \citet{strubell-etal-2019-energy} raised concerns about the environmental impact of training deep learning language models. \citet{patterson_carbon_2021} pointed out that most of the energy consumption for deep learning language models comes during the initial pre-training. In this work, we limit ourselves to fine-tuning and light continued pre-training of T5 and GPT-Neo. While we do not have information about GPT-Neo's training, T5's training took place in highly efficient data centers whose energy consumption was offset by purchasing electricity from renewable sources~\cite{patterson_carbon_2021}. For our light pre-training and fine-tuning, we use a machine with four NVIDIA Quadro RTX 8000 fed from California's energy grid. The total computation time for the experiments in this paper is about 500 hours, but this is an informal estimate rather than an accurate measurement.

\section*{Acknowledgment}
\label{sec:acknowledgment}
This material is based on research supported by DARPA under agreement number N66001-19-2-4032. The U.S. Government is authorized to reproduce and distribute reprints for Governmental purposes notwithstanding any copyright notation thereon. The views and conclusions contained herein are those of the authors and should not be interpreted as necessarily representing the official policies or endorsements, either expressed or implied, of DARPA or the U.S. Government.

\bibliography{main}
\bibliographystyle{acl_natbib}

\clearpage
\appendix

\section{Sample Statements}
\label{sec:sample_statements}
Sample statements in \LEFTS. The first two statements are from \americangovernment, the following three from \ushistory:
\begin{itemize}
    \item Public goods are available to all without payment.
    \item In a majoritarian voting electoral system voters select the party of their choice rather than an individual candidate.
    \item Europeans did not introduce Indians to wampum.
    \item Philadelphia served as the base for British operations for most of the Revolutionary War.
    \item The British bombardment of Baltimore inspired The Star-Spangled Banner.
\end{itemize}
\section{Training Details}
\label{sec:training_details}

For all light pre-training and fine-tuning, we use a machine with four NVIDIA Quadro RTX 8000 GPUs.

\subsection{T5-3B}
\label{sec:t5_hyperparameters}
We implement the model using PyTorch Lightning~\cite{falcon2019pytorch} and Hugging Face's PyTorch Transformers~\cite{wolf-etal-2020-transformers}. For pre-training and fine-tuning, we use a maximum sequence length of 128. We searched for the best learning rate for our model out of $\lbrace 3\mathrm{e}{-5}, 1\mathrm{e}{-4}, 3\mathrm{e}{-4}, 1\mathrm{e}{-3}\rbrace$.

\begin{table}[H]
	\centering
	\scalebox{0.82}{
		\begin{tabular}{rrr}
			& Fine-tuning & Pre-training \\
            \midrule
            Batch size & 16 & 16 \\
    Gradient accumulation & 1 & 1 \\
    Learning rate & 3e-4 & 1e-3 \\
    \midrule
    Num epochs & \multicolumn{2}{c}{20} \\
    Optimizer & \multicolumn{2}{c}{AdamW} \\
    $\beta1$ & \multicolumn{2}{c}{0.9} \\
    $\beta2$ & \multicolumn{2}{c}{0.999} \\
    $\epsilon$ & \multicolumn{2}{c}{1e-8} \\
    Weight decay & \multicolumn{2}{c}{0.0} \\
    Scheduler & \multicolumn{2}{c}{WarmupDecayLR} \\
    Warmup max steps & \multicolumn{2}{c}{400} \\
    fp16 & \multicolumn{2}{c}{no} \\
			\bottomrule
		\end{tabular}
	} 
\label{tbl:t5_hyper}
\caption{Hyperparameters for T5-3B.}
\end{table}

\subsection{GPT-Neo 2.7B}
We use GPT-Neo 2.7B from the Hugging Face Model Hub.\footnote{\url{https://huggingface.co/EleutherAI/gpt-neo-2.7B}} GPT-Neo matches the architecture of GPT-3~\cite{Brown2020}, but is trained on the openly available Pile corpus~\cite{gaoPile800GBDataset}.

\begin{table}[H]
	\centering
	\scalebox{0.82}{
		\begin{tabular}{rrr}
			& Fine-tuning & Pre-training \\
            \midrule
            Batch size & 48 & 2 \\
    Gradient accumulation & 1 & 4 \\
    \midrule
    Num epochs & \multicolumn{2}{c}{10} \\
    Optimizer & \multicolumn{2}{c}{AdamW} \\
    $\beta1$ & \multicolumn{2}{c}{0.9} \\
    $\beta2$ & \multicolumn{2}{c}{0.999} \\
    $\epsilon$ & \multicolumn{2}{c}{1e-8} \\
    Weight decay & \multicolumn{2}{c}{0.01} \\
    Scheduler & \multicolumn{2}{c}{WarmupDecayLR} \\
    Warmup max steps & \multicolumn{2}{c}{200} \\
    fp16 & \multicolumn{2}{c}{yes} \\
			\bottomrule
		\end{tabular}
	} 
\label{tbl:gptneo_hyper}
\caption{Hyperparameters for GPT-Neo.}
\end{table}

\section{Ensuring Statement Correctness}
\label{sec:statement_correctness}

We took several steps to ensure statements' true/false correctness and prevent data bias/tells. For true/false correctness, we manually inspected the statements to check that they correspond to the correct and incorrect choices as given by each textbook's instructor material. We then wrote a script to automatically count the statements for each chapter to ensure that there are as many true labels as there are false. If some labels were to change accidentally during our research, the script would detect the change. For the manually retrieved relevant passages, the humans read each statement and identified the relevant paragraph. In the process, they also checked each statement's label.

To prevent data bias, we wrote statement pairs to have as much word overlap as logically and grammatically possible. We used multiple annotators to write the statements for the two textbooks (two native speakers for \ushistory; one native, one fluent non-native for \americangovernment). No partition is composed of statements written exclusively by a single person, ensuring no person-speciﬁc tells. Following that, we checked all statements for grammar and punctuation issues using automated checkers and another annotator reading. This stage deals with copy-paste tells in the data and cases where statements for one label sound unnatural.

\end{document}